\documentclass[times,twocolumn,final,number]{elsarticle}

\usepackage{prletters}
\usepackage{framed,multirow}

\usepackage{amssymb}
\usepackage{bm}
\usepackage{latexsym}
\usepackage{mathtools} %loads amsmath as well, for matrix
\usepackage{graphicx}
\usepackage[ruled, linesnumbered]{algorithm2e}
\usepackage{adjustbox}
\usepackage{booktabs}     
\usepackage{float}
\usepackage{subfig}
\usepackage{enumitem}
\usepackage{url}
\usepackage{xcolor}
\definecolor{newcolor}{rgb}{.8,.349,.1}

\def\argmax{\mathop{\rm arg\,max}\limits}

\journal{Pattern Recognition Letters}

\begin{document}

\clearpage
\thispagestyle{empty}
\ifpreprint
  \vspace*{-1pc}
\fi

\clearpage

\ifpreprint
  \setcounter{page}{1}
\else
  \setcounter{page}{1}
\fi

\begin{frontmatter}

\title{Perception Matters: Exploring Imperceptible and Transferable Anti-forensics for GAN-generated Fake Face Imagery Detection}

\author[1]{Yongwei {Wang}\corref{cor1}} 
\cortext[cor1]{Corresponding author: 
  Tel.: +1-778-885-5268;  }
\ead{yongweiw@ece.ubc.ca}
\author[2]{Xin {Ding}}
\author[1,3]{Li  {Ding}}
\author[1]{Rabab  {Ward}}
\author[1]{Z. Jane  {Wang}}

\address[1]{Department of Electrical and Computer Engineering, University of British Columbia, Vancouver, V6T1Z4, Canada}
\address[2]{Department of Statistics, University of British Columbia, Vancouver, V6T1Z4, Canada}
\address[3]{School of Information and Communications Engineering, Xi'an Jiaotong University, Xi'an, 710048, China}

\begin{abstract}
Recently, generative adversarial networks (GANs) can generate photo-realistic fake facial images which are perceptually indistinguishable from real face photos, promoting research on fake face detection. Though fake face forensics can achieve high detection accuracy, their anti-forensic counterparts are less investigated. Here we explore more \textit{imperceptible} and \textit{transferable} anti-forensics for fake face imagery detection based on adversarial attacks. Since facial and background regions are often smooth, even small perturbation could cause noticeable perceptual impairment in fake face images. Therefore it makes existing adversarial attacks ineffective as an anti-forensic method. Our perturbation analysis reveals the intuitive reason of the perceptual degradation issue when directly applying existing attacks. We then propose a novel adversarial attack method, better suitable for image anti-forensics, in the transformed color domain by considering visual perception. Simple yet effective, the proposed method can fool both deep learning and non-deep learning based forensic detectors, achieving higher attack success rate and significantly improved visual quality. Specially, when adversaries consider imperceptibility as a constraint, the proposed anti-forensic method can improve the average attack success rate by around 30\% on fake face images over two baseline attacks. \textit{More imperceptible} and \textit{more transferable}, the proposed method raises new security concerns to fake face imagery detection. We have released our code for public use, and hopefully the proposed method can be further explored in related forensic applications as an anti-forensic benchmark.                           
\end{abstract}


\end{frontmatter}

%% main text
\section{Introduction}
\label{sec:intro}
\begin{figure}[htp]
\centering
\includegraphics[width=0.45\textwidth]{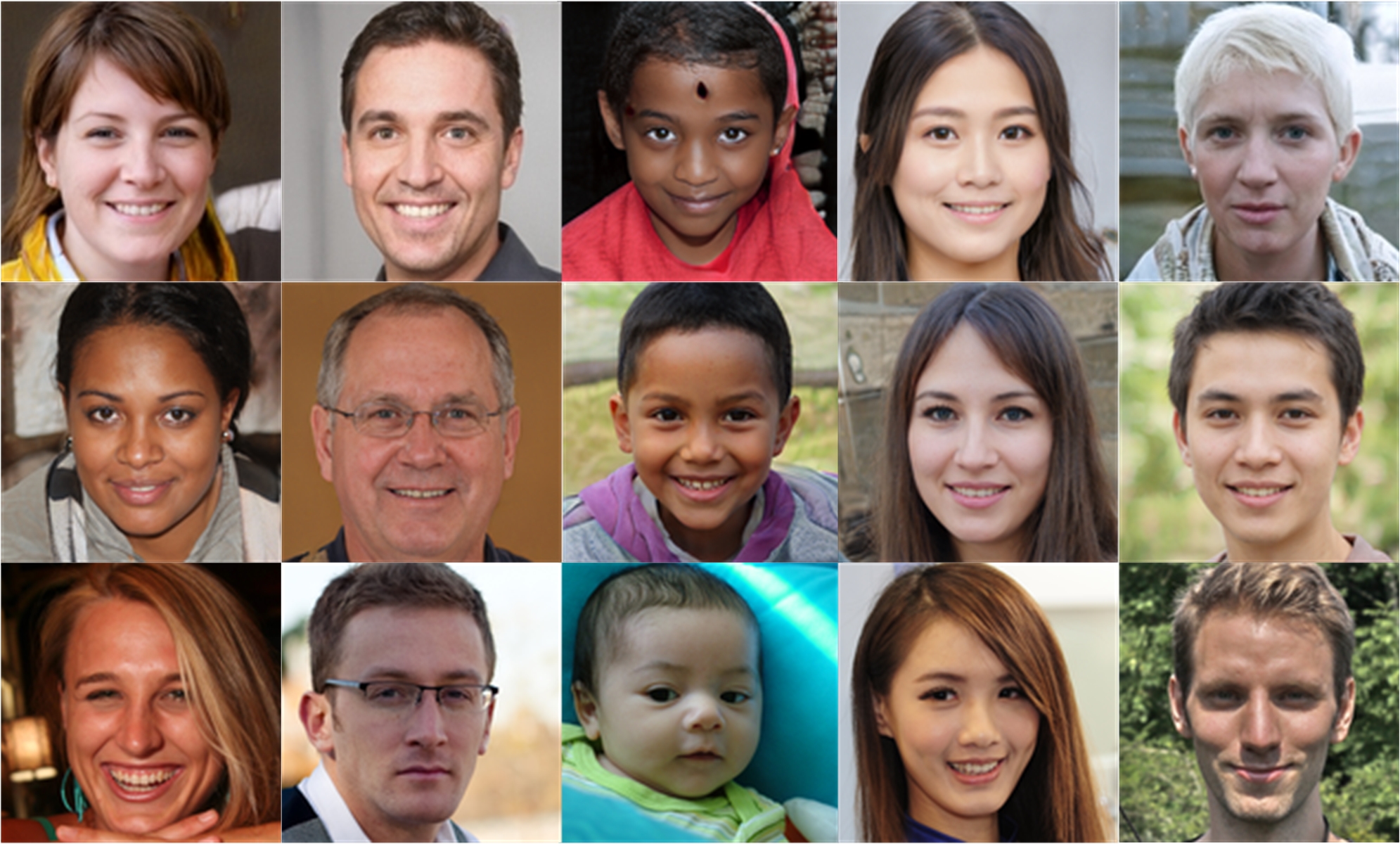}
\centering
\caption{Example images for fake face imagery detection. Question: Which images are from real persons and which ones are generated from GAN? Image samples are from \citep{styleGAN,styleGAN2}.}
\label{fig:fakeFaces}
\end{figure}

Deep neural networks (DNNs) have been playing an overwhelming role in transforming our perspectives towards the digital world \citep{AlexNet,ResNet,SPM18,Revhashnet}. Apart from performing the human-aiding tasks, DNNs can also generate new digital objects/images. Recently generative adversarial networks (GANs) were used to generate photo-realistic fake face photos to easily fool human eyes \citep{GAN14,proGAN,styleGAN,styleGAN2}. In Fig. \ref{fig:fakeFaces}, we show several human face images where some are captured from real person and some are generated from advanced GANs. Can you pick up GAN-generated fake face photos in Fig. \ref{fig:fakeFaces}? (\textit{Answer}: Images in the first two rows are fake face images from styleGAN \citep{styleGAN} and styleGAN2 \citep{styleGAN2}, respectively; while those in the last row are real ones from the Flicker face dataset \citep{styleGAN}.)

The widespread of such visually realistic fake face images may pose security concerns \citep{GANforen18,Wface18,fakeNet18}. E.g., the Washington Post reported that some spies created social accounts with AI-generated fake face images to connect with politicians for malicious purposes \citep{news}. Fake face photos may also be used to falsify identity information and create fake news. Therefore accurate and reliable detection of such fake face images is important. 

In work \citep{GANforen18}, the authors fed features from a pretrained VGG network \citep{VGG} to steganalysis classifiers \citep{steg12} to identify fake face images from real ones. In work \citep{marra2019gans,yu2019attributing}, the authors studied the existence of GAN fingerprints to distinguish fake images generated from different GAN models. In work \citep{colClue18}, the authors analyzed the structure of GAN architectures and proposed to utilize saturation statistics as features, and the extracted features were classified with a support vector machine.     

In work \citep{fakeNet18}, a deep learning-based forensic detector was designed and gave high average accuracy, i.e., over 98\% on fake face detection. The authors firstly cast color images to the residual domain with high-pass filters. Then a set of convolutional modules were applied for feature extraction and classification. More recently, in \citep{wang2020cnn}, the authors proposed a general fake face detector which was shown to generalize well to detect fake images from unseen GAN models. The authors in \citep{color_disp} investigated discernible color disparities between GAN-generated and real face photos. Then ensemble steganalysis classifiers were employed using features extracted from a third order co-occurrence matrix. Among non-deep learning based methods, the method in \citep{color_disp} achieved superior forensic accuracy on fake face imagery detection. 

While existing forensics can successfully identify GAN-generated fake face images, there exists the concern that fake face imagery detectors might be easily bypassed by anti-forensic methods. Image anti-forensics is a countermeasure of image forensics by manipulating discernible traces to reduce the performance of forensic detectors \citep{forenReview13,pivaReview13}. Existing anti-forensic methods often target at specific forensic detectors, e.g., JPEG compression detection \citep{jpeg_anti14}, which are not directly applicable for our fake face detection task.  

Though rarely investigated yet, studying fake face imagery anti-forensics is meaningful since it exposes possible vulnerability issues of forensic detectors. In turn, the anti-forensic study promotes researchers to propose more reliable and robust detectors, which is critical in safety-related forensic tasks. In this study, our contributions are summarized as follows:

1. We introduce adversarial attacks as an automatic anti-forensic approach for GAN-generated fake face detection. Our study shows both deep-learning and non-deep learning based methods can be vulnerable to such adversarial perturbations. 
     
2. We investigated the perturbation residues of existing forensic models both in the $RGB$ and $YC_bC_r$ domains. Our analysis shows that existing gradient-based attacks display strong correlations for perturbations at $RGB$ channels, while such correlations reduce in the $YC_bC_r$ domain. The perturbation mainly concentrates on the $Y$ component, leading to severe visual distortion effects.          
    
3. We propose a novel adversarial attack algorithm with perception constraints in the $YC_bC_r$ domain. We allocate more perturbation for $C_b$ and $C_r$ channels while less for $Y$. \textit{More imperceptible} and \textit{transferable}, the proposed method significantly improves the visual quality and the attack success rate when compared with baseline attacks. We have released our codes, datasets and more results in Github: \url{https://github.com/enkiwang/Imperceptible-fake-face-antiforensic}. 
    
4. Finally, this study also reveals several interesting observations. For example, perturbations crafted for fake face images are significantly more transferable than those for real face images on all attacks we evaluated, which is worthy of further investigation.   

\section{Related Work}
\label{sec:related}

% \vspace{-5pt}
\textbf{GAN-generated fake face imagery}. GAN was formulated as a two-player game between a generator and a discriminator \citep{GAN14,DCGAN}. In theory, the generator can generate visually realistic images by capturing the underlying distributions of real data when GAN reaches an equilibrium. In practice, vanilla GAN models often suffer from training instability issues. Subsequent studies then tried to stabilize training GANs (e.g. \citep{miyato2018spectral,zhang2019self,durall2020watch}). Specific to fake face imagery generation, progressive GAN (ProGAN) \citep{proGAN} was the first GAN model to generate high resolution fake face images with relatively good visual quality. Then Karras et al. developped StyleGAN \citep{styleGAN} which can generate human face photos with impressively realistic visual quality. Recently, StyleGAN2 \citep{styleGAN2} was proposed to achieve the state-of-the-art performance in fake face generation.

\textbf{Adversarial attacks}. Recent studies show DNNs are vulnerable to adversarial perturbations, termed as \textit{adversarial examples} \citep{opt14,FGSM,blackBox1,madry2017towards,MIFGSM}. Adversarial examples crafted from one network can possibly fool an unknown model. This makes adversarial examples as potential threats to deployed safety-critical systems built on DNNs.  Despite active studies in the computer vision area, the existence of adversarial examples has raised relatively less attention in the forensic community \citep{marra2018vulnerability,transfer_icassp19}, which requires forensic detection to be both accurate and secure. For instance, a fake face image detection model is potentially meaningless if it is susceptible to certain carefully crafted perturbation.     

Compared with general adversarial attacks, the anti-forensic method for GAN-generated fake face imagery detection has its unique characteristics. Generally, a higher perturbation budget indicates stronger attack ability, but degradation in visual quality. For natural-scene texture-rich images, relative higher perturbation does not seriously impair the perceptual quality. However, in fake face imagery anti-forensics, facial images are very sensitive to adversarial perturbation due to their large smooth regions. To avoid being spotted, the crafted perturbation should look \textit{\textbf{imperceptible}} to human eyes. Otherwise, such perturbed images can be easily detected by visual sanity check.     

For adversaries, another desirable property is that anti-forensic manipulations are \textit{\textbf{transferable}} to unseen forensic models. \textit{Transferability} means the anti-forensic perturbation designed for specific forensic models can also reduce the detectability of other unknown forensic models. This property also poses severe threats to fake fake forensic detectors.

In work \citep{marra2018vulnerability}, the authors employed existing attack methods \citep{FGSM,madry2017towards} to study the adversarial vulnerability of deep learning-based classifiers for camera model identification. In work \citep{transfer_icassp19}, the authors examined adversarial attacks in the median filtering and image resizing forensic tasks, and concluded that adversarial examples are generally not transferable in image forensics. However, such conventional attack methods they used are less transferable and lead to perceptual issues in our specific anti-forensic task. Therefore, in this study we propose a novel perception-aware attack method which provides both imperceptible visual quality and higher transferability than those from the existing methods.

\section{Method}
\label{sec:antiForensic}

\subsection{The adversarial attack problem}
Assume a forensic detector $f:\mathcal{D} \subseteq \mathbb{R}^d \mapsto \mathbb{R}^K$, where $\mathcal{D}=[0, 255]^d$. Given a data sample $\boldsymbol{x} \in \mathbb{R}^d$, the detector correctly predicts its label as $y \in \mathcal{Y}$, i.e., $y=\argmax_{k=1,\cdots,K} f_k(\boldsymbol{x})$. 

The adversarial attack problem seeks an $\epsilon$-ball bounded perturbation $||\boldsymbol{\delta}||_p \leq \epsilon$ within the vicinity of $\boldsymbol{x}$, which makes the forensic detector fail with a high probability. Here $||\cdot||_p$ denotes the $\ell_p$ norm constraint. Then the perturbed data $\boldsymbol{x}^{adv}:= \boldsymbol{x} + \boldsymbol{\delta} $ is an adversarial example w.r.t the threat model if the following conditions are satisfied,
\begin{equation}
    \argmax_{k=1,\cdots,K} \; f_k(\boldsymbol{x} + \boldsymbol{\delta}) \neq y, \quad ||\boldsymbol{\delta}||_p \leq \epsilon \quad \textrm{and}\quad \boldsymbol{x} + \boldsymbol{\delta} \in \mathcal{D} 
\end{equation}

Denoting a surrogate function as $\mathcal{L}$, we define the constrained optimization problem as,
\begin{equation}
    \argmax_{\boldsymbol{\delta}} \; \mathcal{L}(f(\boldsymbol{x}+\boldsymbol{\delta}), y) \quad \textrm{s.t.} \; ||\boldsymbol{\delta}||_p \leq \epsilon, \; \boldsymbol{x} + \boldsymbol{\delta} \in \mathcal{D}
    \label{eq:obj_fun}
\end{equation}

In this work, we use the $\ell_\infty$ norm constraint, a popular $\ell_p$ norm in the literature. The surrogate function $\mathcal{L}$ is selected as the binary cross entropy function in our setting.  

To solve Eq.(\ref{eq:obj_fun}), \citep{FGSM} proposed the Fast Sign Gradient Method (FGSM), a one-step gradient-based perturbation, which utilizes the sign of the gradient w.r.t. the input data,
\begin{equation}
    \boldsymbol{\delta}_{FGSM} = \epsilon \cdot \textrm{sign}(\nabla_{\boldsymbol{x}} \mathcal{L} (f(\boldsymbol{x}+\boldsymbol{\delta}, y))
    \label{eq:fgsm}
\end{equation}
where the element-wise $\textrm{sign}(\cdot)$ function gives $+1$ for positive values, and $-1$ for negative values; otherwise, it gives $0$. 

The FGSM method was designed under the assumption that the decision boundary is linear around the input data. For neural networks with nonlinear activation function, this assumption does not hold, thus the FGSM attack generally ``underfits'' the model, which compromises its attack ability. To increase the attack ability, adversaries can apply Eq.(\ref{eq:fgsm}) iteratively for multiple times \citep{madry2017towards}. \citep{MIFGSM} further incorporates the momentum during the gradient update at each iteration and proposes the Momentum Iterative-FGSM (MIM). We use FGSM (single-step) and MIM (multiple-step) as our baseline attacks.  

\subsection{Perturbation analysis in $YC_bC_r$ domain}
In this section, we investigate spatial correlations of adversarial perturbations in $R$, $G$, and $B$ channels. We then show that for existing fake face forensic models (trained on RGB domain), with baseline attacks, the perturbation energy concentrates more in the $Y$ component than in $C_b$ and $C_r$ components. 

For simplicity, we analyze adversarial perturbations generated from FGSM, the single-step attack method with perturbation as the gradient (after sign). For a single pixel in an image, we denote the gradient (after sign) of $R$, $G$, $B$ components as three random variables $\boldsymbol{S} = (s^r, s^g, s^b)^T$, where $s^r, s^g, s^b$ follows the Bernoulli distribution. The statistical correlations of these three components are provided by the covariance matrix $\Sigma_{\boldsymbol{S}}$, which can be estimated via observations of the random variable $\boldsymbol{S}$,
\begin{equation}
    \boldsymbol{\Sigma_{\boldsymbol{S}}} \approx \frac{1}{N} \sum_{i=1}^{N} (\boldsymbol{S}_i - \bar{\boldsymbol{S}}) \cdot (\boldsymbol{S}_i - \bar{\boldsymbol{S}})^T
    \label{eq:cov_mat}
\end{equation}
where $N$ denotes the number of observations of $\boldsymbol{S}$, and $\bar{\boldsymbol{S}}$ represents the sample mean of $\boldsymbol{S}$.  
The conversion from the $RGB$ domain to the $YC_bC_r$ domain is to perform an affine transformation, 
\begin{equation}
    \boldsymbol{S}' = \boldsymbol{A} \boldsymbol{S} + \boldsymbol{b}
    \label{eq:ycbcr}
\end{equation}
where $\boldsymbol{S}'=(s^y, s^{Cb}, s^{Cr})^T$ denotes the transformed random variables in the $YC_bC_r$ domain; $\boldsymbol{A}, \boldsymbol{b}$ denote respectively the transformation matrix and bias, with
\[\boldsymbol{A} = \left[ 
\begin{matrix}
0.2568 & 0.5041 & 0.0979 \\
    -0.1482 & -0.2910 & 0.4392 \\
    0.4392 & -0.3678 & -0.0714 \\
\end{matrix}
\right]
\]
and \[
\boldsymbol{b} = \left(16, 128, 128 \right)^T
\]

Then we can obtain the covariance matrix of $\boldsymbol{S}'$ as,
\begin{equation}
    \Sigma_{\boldsymbol{S}'} = \boldsymbol{A} \boldsymbol{\Sigma_{\boldsymbol{S}}} \boldsymbol{A}^T
\end{equation}

\begin{figure}[h]
\centering
\includegraphics[width=0.45\textwidth]{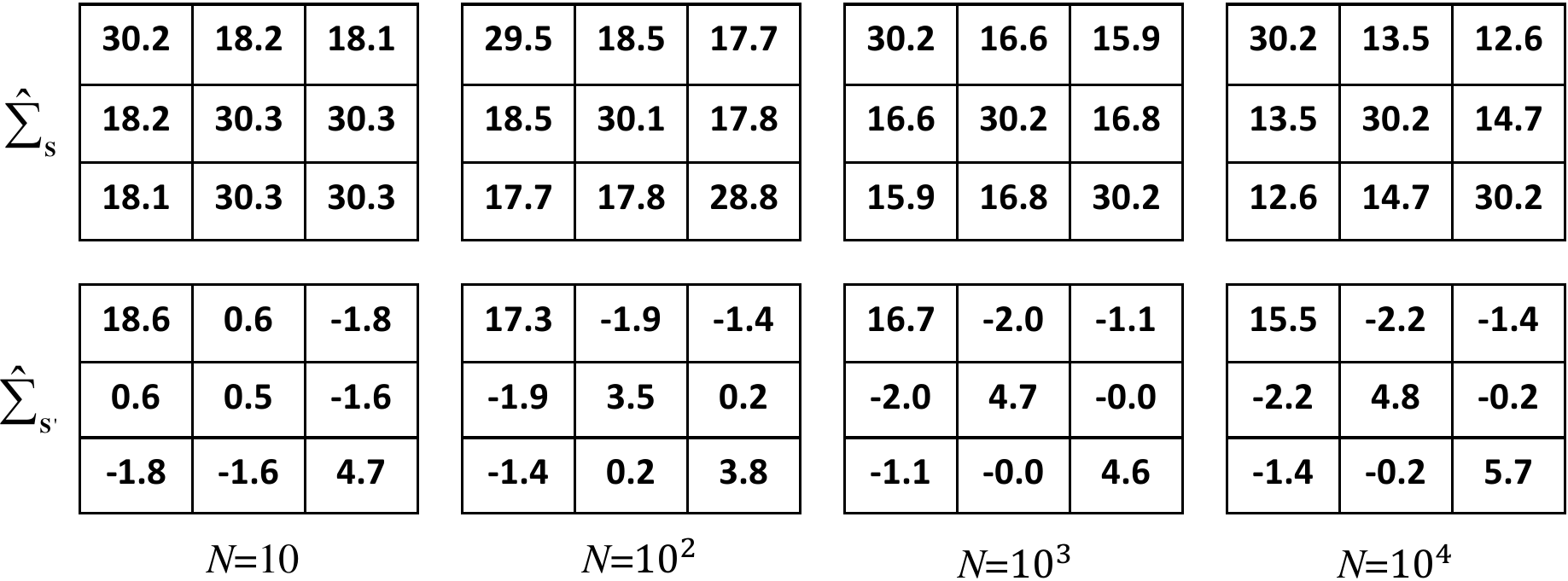} 
\caption{Illustration of estimated covariance matrices $\hat{\Sigma}_{\boldsymbol{S}}$ ($1^{st}$ row) and $\hat{\Sigma}_{\boldsymbol{S}'}$ ($2^{nd}$ row) with  $N=10,10^2,10^3$ and $10^4$ respectively. Here $\epsilon=5.5$.} 
\label{fig:cov_mat}
\end{figure}

In Fig. \ref{fig:cov_mat}, we illustrate the covariance matrices of $\boldsymbol{S}$ and $\boldsymbol{S}'$ estimated with the number of pixels as $N=10,10^2,10^3$ and $10^4$ on StyleGAN \citep{styleGAN}. Clearly, $s^r, s^g, s^b$ components are highly correlated; while the correlations reduce when we apply the $YC_bC_r$ transform in Eq.(\ref{eq:ycbcr}). Also, we notice that the variances are almost identical for $s^r, s^g, s^b$, while the variance of $s^y$ is significantly larger than that of $s^{Cb}$ and $s^{Cr}$ (i.e., around $3$ times larger). It indicates that the perturbation energy concentrates more on the $Y$ component than on $C_b$ and $C_r$ components. 

\begin{figure}[h]
\centering
\includegraphics[width=8.2cm, height=4.5cm]{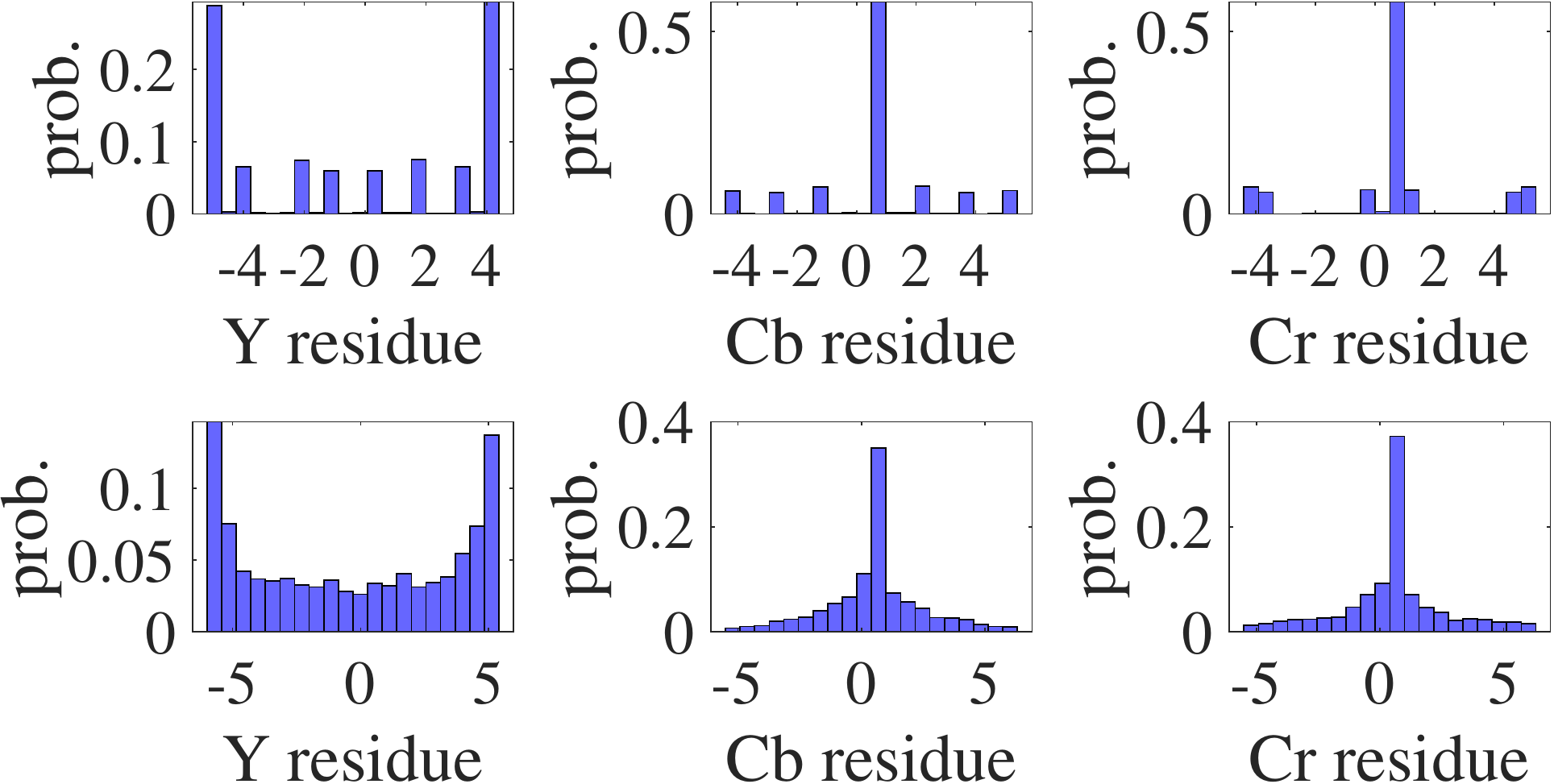} 
\caption{Example perturbation histograms of FGSM ($1^{st}$ row) and MIM ($2^{nd}$ row) attacks in the $YC_bC_r$ domain. The histogram is generated by using 5000 adversarial samples of StyleGAN-generated fake face images. The perturbation bounds are $\epsilon=5.5$ and $\epsilon=6$ for FGSM and MIM, respectively.}
\label{fig:res_fgsm_mim}
\end{figure}

To validate the analysis, we generate adversarial examples using FGSM and MIM, and show the histograms of perturbations in the $YC_bC_r$ domain in Fig. \ref{fig:res_fgsm_mim}. For both attacks, we observe that perturbation residues mainly cluster at $\pm 5.5$ for $Y$ while the perturbations peak around $0$ for $C_b$ and $C_r$ components. We observed similar perturbation phenomena during attacking existing forensic models on StyleGAN2 \citep{styleGAN2} and ProGAN \citep{proGAN} datasets. Since the human visual system is more sensitive to perturbations in the $Y$ component than in $C_b$ and $C_r$ components, this intuitively explains why the $RGB$ domain attacks are prone to visual distortion.  

\vspace{-3mm}
\subsection{Proposed adversarial attack}
Based on the perturbation analysis discussed above, as an alternative to existing attacks on the RGB domain, we are motivated to perform adversarial attacks with explicit perturbation constraints in the $YC_bC_r$ domain. By exploiting the perception characteristics, we propose to directly allocate more perturbations to $C_b$ and $C_r$ components than to the $Y$ component to produce more visually pleasant adversarial examples.

Denote $\mathcal{T}$ as the transformation operator from the $RGB$ domain to the $YC_bC_r$ domain (see Eq.(\ref{eq:ycbcr})), and $\mathcal{T}^{-1}$ as its inverse transformation back to the $RGB$ domain. The proposed loss function is expressed as,

\begin{equation}
\begin{split}
    \mathcal{L}\left( f \left(\mathcal{T}^{-1} (\mathcal{T} \boldsymbol{x} + \boldsymbol{ \zeta}) \right), y \right) = & -y \cdot \textrm{log} \left(f_y\left(\mathcal{T}^{-1} (\mathcal{T} \boldsymbol{x} + \boldsymbol{ \zeta}) \right) \right) \\
    & - (1-y) \cdot \textrm{log} \left( f_{1-y} \left(\mathcal{T}^{-1} (\mathcal{T} \boldsymbol{x} + \boldsymbol{ \zeta}) \right) \right)
\end{split}
\label{eq:prop_loss}
\end{equation}
where $\boldsymbol{\zeta}$ denotes the perturbation that is directly optimized in the $YC_bC_r$ domain. 

Now our constrained optimization problem becomes,
\begin{equation}
\begin{split}
    \argmax_{\boldsymbol{\zeta}} \; & \mathcal{L}\left( f \left(\mathcal{T}^{-1} (\mathcal{T} \boldsymbol{x} + \boldsymbol{ \zeta}) \right), y \right) \\
      & \textrm{s.t.} \; ||\boldsymbol{\zeta}^{[c]}||_{\infty} \leq \epsilon^{[c]}, \; c\in \left\{Y, C_b, C_r \right\} , \\
      & \textrm{and} \quad \boldsymbol{x} + \mathcal{T}^{-1} \boldsymbol{\zeta} \in \mathcal{D}
    \label{eq:obj_fun_prop}
\end{split}
\end{equation}
where $\boldsymbol{\zeta}^{[c]}$ and $\epsilon^{[c]}$ denote the constrained perturbation and its perturbation budget at channel $c, c\in \{Y, C_b, C_r \}$, respectively. To alleviate the visual distortion effects due to perturbations $\boldsymbol{\zeta}$, it is desirable to assign larger values for $\epsilon^{[C_b]}, \epsilon^{[C_r]}$ than $\epsilon^{[Y]}$. Assume that we have access to the forensic detector (or its substitute model), we can utilize the gradient-based approach to solve Eq.(\ref{eq:obj_fun_prop}). 

Denote any pixel in an image by $\boldsymbol{P}_{i,j}=\left( R(i,j), G(i,j), B(i,j) \right)^T$, and its counterpart in the $YC_bC_r$ domain as $\boldsymbol{P}_{i,j}'=\left( Y{(i,j)}, C_b{(i,j)}, C_r{(i,j)} \right)^T$. We can propagate the gradient from the $RGB$ to the $YC_bC_r$ domain,
\begin{equation}
\begin{split}
    \nabla_{\boldsymbol{P}_{i,j}'} \mathcal{L}\left( f \left(\mathcal{T}^{-1} (\mathcal{T} \boldsymbol{x} + \boldsymbol{ \zeta}) \right), y \right) &= \left( \boldsymbol{1} \oslash \boldsymbol{A}  \right) \; \cdot  \\
    & \nabla_{\boldsymbol{P}_{i,j}} \mathcal{L} \left( f \left(\mathcal{T}^{-1} (\mathcal{T} \boldsymbol{x} + \boldsymbol{ \zeta}) \right), y \right)
\end{split}
\label{eq:grad_cal}
\end{equation}
where $\nabla_{\boldsymbol{P}_{i,j}'} \mathcal{L}= \left(\frac{\partial \mathcal{L}}{\partial Y(i,j)}, \frac{\partial \mathcal{L}}{\partial C_b(i,j)}, \frac{\partial \mathcal{L}}{\partial C_r(i,j)} \right)^T$ and $\nabla_{\boldsymbol{P}_{i,j}} \mathcal{L}= \left(\frac{\partial \mathcal{L}}{\partial R(i,j)}, \frac{\partial \mathcal{L}}{\partial G(i,j)}, \frac{\partial \mathcal{L}}{\partial B(i,j)} \right)^T$ denote the partial derivatives w.r.t. the loss function $\mathcal{L}(\cdot)$ in $RGB$ and $YC_bC_r$ domains, respectively; $\oslash$ denotes the elementwise division operation. 

The flowchart of the proposed attack method is described in detail in Algorithm \ref{alg:ycc_prop}.   

\begin{algorithm}[ht]
	\footnotesize
	\SetAlgoLined
	\KwData{A clean image $\boldsymbol{x}$ with label $y$, a fake-face forensic model $f$, channel-wise perturbation budget $\epsilon^{[c]}, c\in\{Y, C_b, C_r\}$, iteration number $K$ and hyperparameter $\mu$.} 
 	\KwResult{Optimized perturbation $\boldsymbol{\zeta}$ that satisfies $\left\{\boldsymbol{\zeta} \; | \; \{||\boldsymbol{\zeta}^{[c]}||_{\infty} \leq \epsilon^{[c]} \} \cap \{\boldsymbol{x} + \mathcal{T}^{-1} \boldsymbol{\zeta} \in \mathcal{D} \} \right\}$, and the perturbed image  $\boldsymbol{x}^{adv}$.}   
	Initialize $\alpha^{[c]}={\epsilon^{[c]}}/{K}, c\in\{ Y, C_b, C_r \}$, $\boldsymbol{\zeta}_{(0)}=\boldsymbol{0}$, $\boldsymbol{g}_{(0)}'=\boldsymbol{0}$\;
	\For{$k=0$ \KwTo $K-1$}{
	Input $\boldsymbol{x}_{(k)}$ to the forensic model $f$, and compute gradients of $\boldsymbol{x}$:  $\nabla_{\boldsymbol{x}_{(k)}} \mathcal{L}$\;
	Compute gradients w.r.t. $\mathcal{T}\boldsymbol{x}_{(k)}$ using Eq.(\ref{eq:grad_cal}): $\nabla_{\mathcal{T}\boldsymbol{x}_{(k)}} \mathcal{L}$\;
	Compute accumulated gradients w.r.t. $\mathcal{T}\boldsymbol{x}_{(k)}$: 
	$\boldsymbol{g}_{(k+1)}'=\mu \cdot \boldsymbol{g}_{(k)}' + \nabla_{\mathcal{T}\boldsymbol{x}_{(k)}} \mathcal{L} / ||\nabla_{\mathcal{T}\boldsymbol{x}_{(k)}} \mathcal{L}||_1$\;
	Compute perturbation $\boldsymbol{\zeta}_{(k+1)}$: $\boldsymbol{\zeta}^{[c]}_{(k+1)}=\boldsymbol{\zeta}^{[c]}_{(k)}+\alpha^{[c]} \cdot \textrm{sign} \left( \boldsymbol{g}_{(k+1)}' \right), c\in \{Y, C_b, C_r \}$\;
	Project $\boldsymbol{\zeta}_{(k+1)}$ within the $\epsilon$-ball: $\boldsymbol{\zeta}_{(k+1)}=\textrm{max} \left( \textrm{min} \left(\boldsymbol{\zeta}_{(k+1)}, \epsilon \right), -\epsilon \right)$\;
	Update adversarial example $\boldsymbol{x}_{(k+1)}$: $\boldsymbol{x}_{(k+1)}= \boldsymbol{x} + \mathcal{T}^{-1}\boldsymbol{\zeta}_{(k+1)} $\;
	Project $\boldsymbol{x}_{(k+1)}$within the feasible set $\mathcal{D}$: $\boldsymbol{x}_{(k+1)}=\textrm{Proj}_{\mathcal{D}} \left( \boldsymbol{x}_{(k+1)} \right)$\;
	}
	\textbf{Return}: Optimized perturbation $\boldsymbol{\zeta}=\boldsymbol{\zeta}_{(K)}$ and the perturbed image $\boldsymbol{x}^{adv} = \boldsymbol{x}_{(K)}$.
	\caption{The proposed algorithm of adversarial attacks in the $YC_bC_r$ domain.}
	\label{alg:ycc_prop}
\end{algorithm}

\vspace{-3mm}
\section{Experiments}
\label{sec:experiment}
\subsection{Experimental setup}
\textbf{Datasets:} We create face image datasets for the fake face imagery detection task: Dataset 1 and Dataset 2, respectively. Dataset 1 consists of 40,000 real face photos and 40,000 StyleGAN-generated photo-realistic facial images \citep{styleGAN}. In Dataset 2, the fake face images are from StyleGAN2 \citep{styleGAN2}. For real or fake images in both datasets, image splits are: 30,000 images for model training, 5,000 images for validation and the rest 5,000 images for test. To reduce the computational complexity, all images are resized to $128\times128$. 

\textbf{Models:} We study seven effective fake face identification models \citep{fakeNet18,VGG, DCGAN, AlexNet, MobileNetV2, wang2020cnn, color_disp}, which are trained from scratch on the face datasets described above. For deep learning-based models (trained on RGB domain), the hyperparameters are as follows: the learning rate is set as $10^{-4}$ with weight decay $5\times 10^{-4}$, the batchsize is selected as 64, and the number of epochs equals 20 with early stopping. For non-deep learning based fake-face detection models \citep{colClue18,color_disp}, we consider the state-of-the-art method proposed in \citep{color_disp}. For convenient expression, we denote the deep-learning based forensic models as ${m}_i, i=1,2,\cdots, 6$ for six different architecture from work \citep{fakeNet18,VGG, DCGAN, AlexNet, MobileNetV2, wang2020cnn} used in the literature respectively. We denote the selected non-deep learning forensic model as ``\textit{NDL}'' \citep{color_disp}. 

To make sure that the forensic models work well (e.g., detection accuracy $\ge 90\%$), we adopt the true positive rate ($TPR$) and the true negative rate ($TNR$) as their performance measures, where $TPR$ and $TNR$ are defined as:
\begin{equation}
    TPR = \frac{TP}{TP + FN}, \quad TNR = \frac{TN}{TN + FP}
\end{equation}
where $TP$, $TN$, $FP$ and $FN$ denote the numbers of correctly identified fake face images, correctly detected real face images, misclassified real face samples and misclassified fake face images, respectively. A good detector provides high $TPR$ and $TNR$ simultaneously. After proper training, all forensic models achieve high $TPR$ and $TNR$ values on both datasets, as shown in Table \ref{tab:model_acc}. We have released these pretrained models to the public.

\begin{table}[htb]
\caption{Pretrained forensic models we evaluated and their performances measured by $TPR$ and $TNR$ on Dataset 1 and Dataset 2, respectively.}
\centering
\begin{adjustbox}{width=0.46\textwidth}
\begin{tabular}{ccccccccc}
\toprule
Datasets                    & models   & $m_1$   & $m_2$   & $m_3$   & $m_4$   & $m_5$   & $m_6$   & $NDL$  \\ \hline 
\multirow{2}{*}{Dataset 1} & TPR (\%) & 98.6 & 94.1 & 91.4 & 95.8 & 90.8 & 99.6 & 98.6 \\ \cline{2-9} 
                           & TNR (\%) & 98.7 & 96.9 & 94.6 & 98.0 & 94.4 & 99.9 & 98.7 \\ \midrule[0.25mm] 
\multirow{2}{*}{Dataset 2} & TPR (\%) & 98.8 & 99.0 & 98.1 & 98.5 & 96.2 & 99.9 & 99.5 \\ \cline{2-9} 
                           & TNR (\%) & 99.2 & 99.4 & 98.5 & 98.5 & 97.2 & 99.9 & 99.4 \\ 
\bottomrule                           
\end{tabular}
\end{adjustbox}
\label{tab:model_acc}
\end{table}

\textbf{Parameters:} In the following experiments, following the baseline MIM method \citep{MIFGSM}, for iterative attacks, we set the iteration number $K$ as 10, and the momentum decay factor $\mu$ as 1. The perturbation bound $\epsilon$ is often chosen as 16. However, this perturbation bound is generally too large in the fake face anti-forensic tasks since it can severely degrade visual quality. To have a good trade-off between visual quality and attack success rate, we set lower perturbation bound, e.g., on Dataset 1 we use $\epsilon$ as $5.5$ and $6$ for FGSM and MIM attacks, respectively. For the proposed method, we set larger values for $\epsilon^{[C_b]}$ and $\epsilon^{[C_r]}$ than $\epsilon^{[Y]}$ for better visual imperceptibility.         

\subsection{Attack success rate comparison}  
The attack success rate ($ASR$) is defined as the accuracy reduction of forensic models after applying adversarial attacks. Concretely, for the fake face detection problem, denote $TPR'$ as the true positive rates after the attack on fake face images. Then $ASR^{[p]}$ on this given fake face image subset (5,000 images in total) is calculated as,
\begin{equation}
    ASR^{[p]} = TPR - TPR'
\label{eq:ASR_tpr}
\end{equation}
Similarly, we can define the attack success rate on real images as $ASR^{[n]} = TNR - TNR'$, where $TNR'$ denotes the true negative rates after the attack on the real face image subset. Clearly, the stronger the adversary, the higher the attack success rates.  

For the visual quality evaluation, we use three popular image quality assessment (IQA) metrics: the ''Naturalness Image Quality Evaluator`` (NIQE) \citep{mittal2012making}, a no-reference IQA to evaluate the naturalness of images (lower indices indicate more natural visual quality); the ''Learned Perceptual Image Patch Similarity'' (LPIPS) \citep{zhang2018unreasonable}, a DL-based IQA for semantic similarity measurement (lower values suggest closer semantic similarity); and the feature similarity index ($\textrm{FSIM}_c$) \citep{zhang2011fsim}, a full-reference IQA based on human visual system (normalized within $[0,1]$, the higher the index, the better the visual quality).   

As an adversary, we focus on attacking fake face images whose reliable detection is vital for forensic models. First, assume we have full access to $m_1$, then we can craft adversarial perturbations based on this model. On Dataset 1, $\epsilon$ are set as 5.5 and 6 for FGSM and MIM attacks, respectively. To have comparable average ASRs, the proposed method adopts $\epsilon^{[Y]}=2.5, \; \epsilon^{[C_b]}=6, \; \epsilon^{[C_r]}=6$. Similarly, on Dataset 2, we use $\epsilon$ as 6 and 7.5 for FGSM and MIM; and $\epsilon^{[Y]}=2, \; \epsilon^{[C_b]}=6, \; \epsilon^{[C_r]}=6$ for the proposed method for a fair comparison. On both datasets, the comparison results are reported in Table \ref{tab:ASR_fake}. The effects of adversarial perturbations crafted from other deep learning models are investigated in Section \ref{sec:transfer}. 

\begin{table*}[htb]
\caption{Performance comparisons of the attack success rate $(\%)$ and the visual quality when applying FGSM, MIM and the proposed method on fake face images from Dataset 1 and Dataset 2. The source model is $m_1$. On Dataset 1, $\epsilon$ is 5.5, 6 for FGSM and MIM attacks respectively; and on Dataset 2, $\epsilon$ is 6, 7.5 for FGSM and MIM, respectively. For the proposed method, $\epsilon^{[c]}$ are $2.5/6/6$ on Dataset 1 and $2/6/6$ on Dataset 2. The best performances are marked in bold.}
\centering
\begin{adjustbox}{width=0.85\textwidth} 
\begin{tabular}{ccccccccccccc}
\toprule
Datasets                   & Attack & $m_1$   & $m_2$   & $m_3$   & $m_4$   & $m_5$   & $m_6$   & $NDL$  & avg. $ASR^{[p]}$ & $\textrm{NIQE}$  & $\textrm{LPIPS}$ & $\textrm{FSIM}_c$ \\ \hline 
\multirow{3}{*}{Dataset 1} & FGSM   & 98.6 & 90.9 & 73.4 & 58.9 & 20.6 & 72.5 & 97.6 & 73.2                                  & 1.188 & 0.026 & 0.955   \\ \cline{2-13} 
                           & MIM    & 98.6 & \textbf{91.6} & 77.4 & 63.4 & 21.1 & \textbf{81.2} & 98.6 & 76.0                                  & 1.032 & 0.028 & 0.952   \\ \cline{2-13} 
                           & \textbf{Prop.}  & \textbf{98.6} & 91.2 & \textbf{83.3} & \textbf{78.3} & \textbf{40.9} & 63.3 & \textbf{98.6} & \textbf{79.2}                                  & \textbf{0.798} & \textbf{0.020} & \textbf{0.984}   \\ \midrule[0.2mm]
\multirow{3}{*}{Dataset 2} & FGSM   & 98.8 & 98.9 & 84.2 & 94.6 & 5.0  & 2.6  & 62.5 & 63.8                                  & 1.728 & 0.034 & 0.969   \\ \cline{2-13} 
                           & MIM    & 98.8 & 99.0 & 97.5 & 97.5 & 6.4  & \textbf{23.3} & 41.0 & 66.2                                  & 1.660 & 0.036 & 0.965   \\ \cline{2-13} 
                           & \textbf{Prop.}  & \textbf{98.8} & \textbf{99.0} & \textbf{98.1} & \textbf{98.5} & \textbf{12.9} & 14.4 & \textbf{92.6} & \textbf{73.5}                                  & \textbf{1.029} & \textbf{0.018} & \textbf{0.992}   \\ 
\bottomrule                           
\end{tabular}
\end{adjustbox}
\label{tab:ASR_fake}
\end{table*}

In Table \ref{tab:ASR_fake}, we can see that with comparable average $ASR^{[p]}$ on both datasets, the perceptual quality of the proposed method has been improved over FGSM and MIM attacks by a large margin quantitatively measured by three IQA metrics. Particularly on Dataset 2, the proposed method achieves considerably improved visual performance and $9.7\%$ and $7.3\%$ higher attack success rates on average on fake face imagery antiforensics.  

\subsection{Visual quality comparison}
As shown in Table \ref{tab:ASR_fake}, when compared with baseline attacks, quantitatively, the proposed method has much improved IQA indices measured by NIQE, LPIPS and $\textrm{FSIM}_c$, i.e., we have higher fidelity with cleaner images either semantically or visually when using the proposed attack algorithm. 

\begin{figure}[h]
\centering
\includegraphics[width=0.48\textwidth]{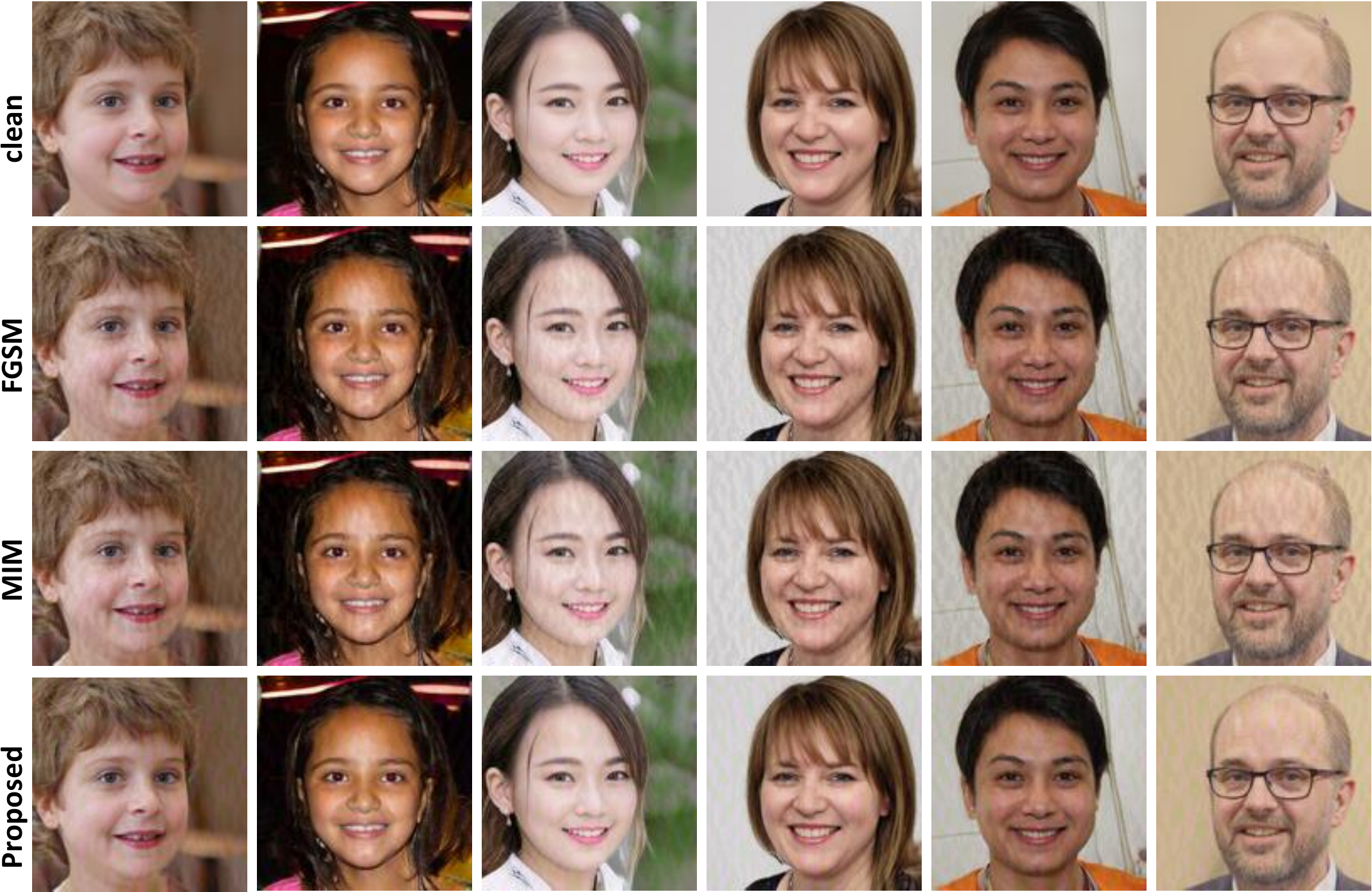} 
\caption{Examples of fake face images for visual quality comparisons on FGSM, MIM and the proposed method. For FGSM and MIM, $\epsilon$ are 5.5 and 6 respectively; for the proposed method, $\epsilon^{[c]}$ are $2.5/6/6$ for $Y, C_b, C_r$ channels. We recommend to zoom in the digital images for better visual comparison. }
\label{fig:vis_fake_face}
\end{figure}

In Fig. \ref{fig:vis_fake_face}, we show several perturbed fake face image examples from Dataset 1. The first row shows clean images, while the rest three rows display their perturbed versions using FGSM, MIM and the proposed method, respectively. By zooming in Fig.\ref{fig:vis_fake_face}, we can easily spot texture-like visual distortions on FGSM and MIM attacks, both in facial regions and background. By contrast, adversarial images from the proposed method still maintain smooth and appear more natural and more \textit{\textbf{imperceptible}}, compared with the clean images. More comparison examples can be found in the project website.

Moreover, we conduct the human subjective preference study to further validate the visual/quantitative comparison results. For each dataset, we randomly choose 50 comparison pairs: clean image and their adversarial version generated by FGSM, MIM and the proposed method, respectively. For each surveyed pair, we prepare two questions: (a) Is it hard to tell which perturbed one is the "cleanest"? If yes, we proceed to the next pair; otherwise we ask the interviewer to (b) Choose the "cleanest" one from three adversarial images. Overall, on each dataset, we received 500 answers (10 volunteers for each dataset), and our subjective study shows that all interviewers perceive adversarial images from the proposed method as the best/cleanest one. With the human preference survey, we can safely conclude that the proposed method has indeed considerably improved the perceptual quality of adversarial images.

\subsection{Adversarial transferability}
\label{sec:transfer}
\begin{figure}[h]
\centering
\subfloat{\includegraphics[width=8.8cm, height=2.7cm]{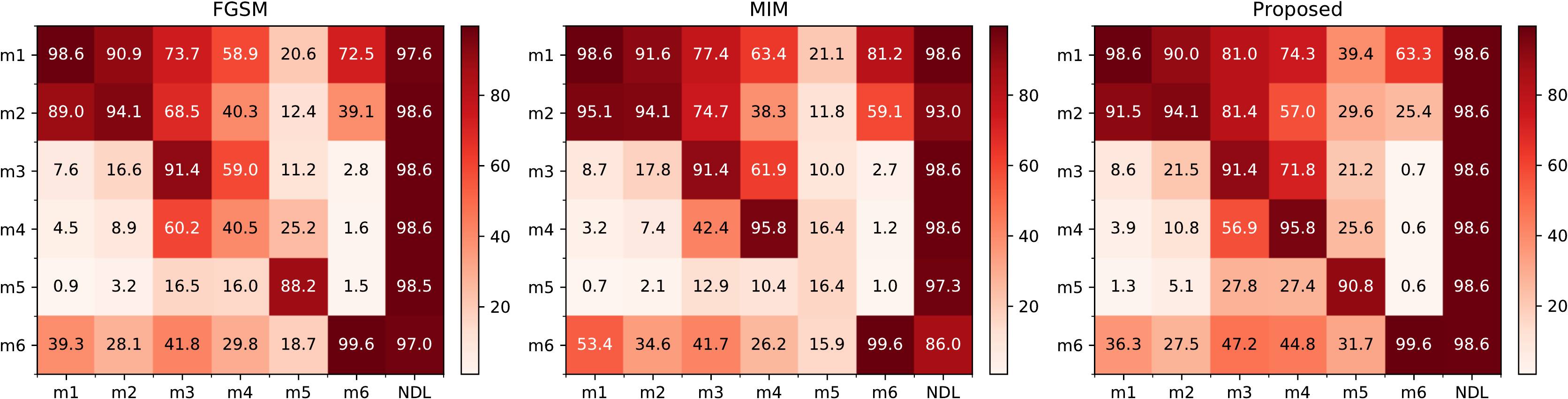}} \\
\subfloat{\includegraphics[width=8.8cm, height=2.7cm]{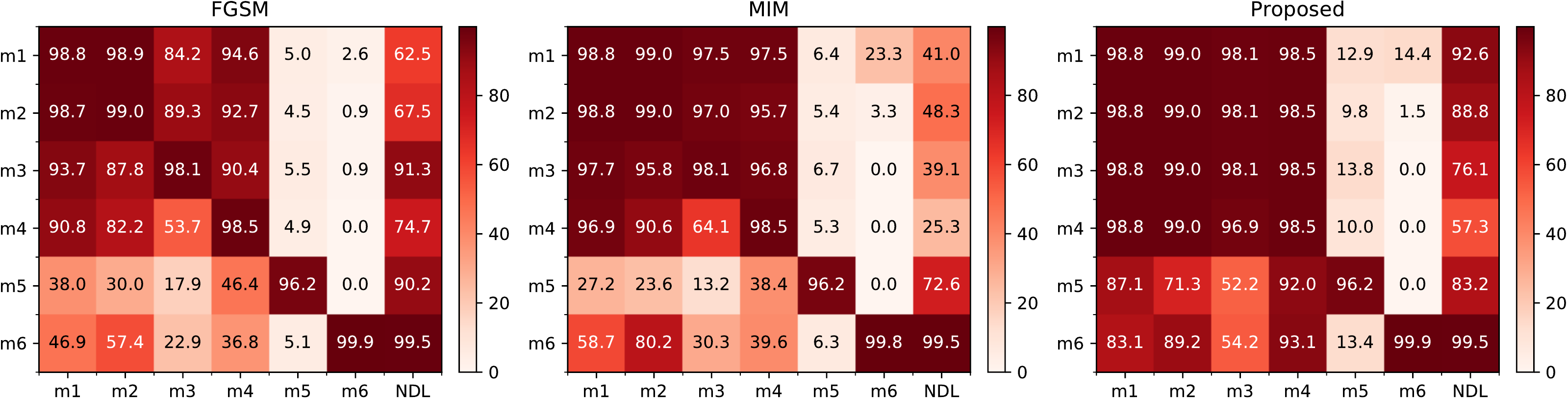}}
\caption{Comparisons of adversarial transferability of FGSM, MIM and the proposed method on fake face image forensic models on Dataset 1 (the $1^{st}$ row) and Dataset 2 (the $2^{nd}$ row), respectively.}
\label{fig:vis_transfer} 
\end{figure}

In Fig.\ref{fig:vis_transfer}, on Dataset 1 and Dataset 2, we visualize the transfer matrices of adversarial examples crafted from different forensic models using FGSM, MIM and the proposed method, respectively. In each matrix, each row denotes the same source model to craft adversarial examples and each column represents a target model on which to be evaluated.

For each source model, the proposed method achieves higher average attack success rates over FGSM and MIM on both datasets. Besides, we have several interesting observations. First, the $NDL$ models are also likely to be fooled in the presence of antiforensic perturbations. Particularly on Dataset 1, $NDL$ models almost completely fail. This indicates that even non-deep learning based forensic models can be vulnerable to adversarial perturbations crafted from deep forensic models, which \textit{necessitates further security investigation into conventional forensic models} in the presence of adversarial attacks. Second, the adversarial transferability can be quite asymmetric between different forensic models. For instance, adversarial perturbations crafted from $m_1$ effectively transfer to $m_4$ for all three attacks. However, adversarial examples created from the source model $m_4$ hardly transfer to $m_1$. This intriguing phenomenon might be related with the sophisticated decision landscapes of DL models (which differ in network modules or depth). We also observe that by the careful selection of source forensic models, \textit{adversaries can build more transferable attacks} with the same attack method. To demonstrate further in this direction, in our preliminary study, we ensemble different forensic models to compose new source models and evaluate their attacking performances. As an example, by using the grid search, we combine three forensic models as $m_{ens(i,j,k)}$ with $i, j, k \in \{1, \cdots, 6 \}$. Then we fuse their scores together with equal weights and generate adversarial perturbations with the proposed method. The average ASRs of some ensemble source models are reported in Table \ref{tab:ASR_ens}. Though it remains unclear on the optimal model ensemble selection (i.e., in terms of the model number and weights), we find some combinations indeed generate more transferable attacks, e.g., the ensemble model $m_{ens(1,4,6)}$ on Dataset 1 and $m_{ens(3,4,5)}$ on Dataset 2. We will investigate further on this phenomenon in the future.

\begin{table}[htb]
\caption{ Average $ASR^{[p]}$ results (\%) from example combinations of the source models on Dataset 1 (\#1) and Dataset 2 (\#2).}
\centering
\begin{adjustbox}{width=0.48\textwidth} 
\begin{tabular}{cccccc}
\toprule
Source model & $m_{ens(1,2,5)}$ & $m_{ens(1,4,5)}$ & $m_{ens(1,4,6)}$ & $m_{ens(2,3,4)}$ & $m_{ens(3,4,5)}$   \\ \hline
avg. $ASR^{[p]}$ (\#1)     &   78.5            &  78.5             &    \textbf{82.2}           &       72.4        &     57.1              \\ \hline
avg. $ASR^{[p]}$ (\#2)     &   83.3            &   83.2            &    72.5           &       72.5        &     \textbf{84.3}               \\ \bottomrule
\end{tabular}
\end{adjustbox}
\label{tab:ASR_ens}
\end{table}

\subsection{Perturbation residues}
\begin{figure}[h]
\centering
\includegraphics[width=8.2cm, height=4.5cm]{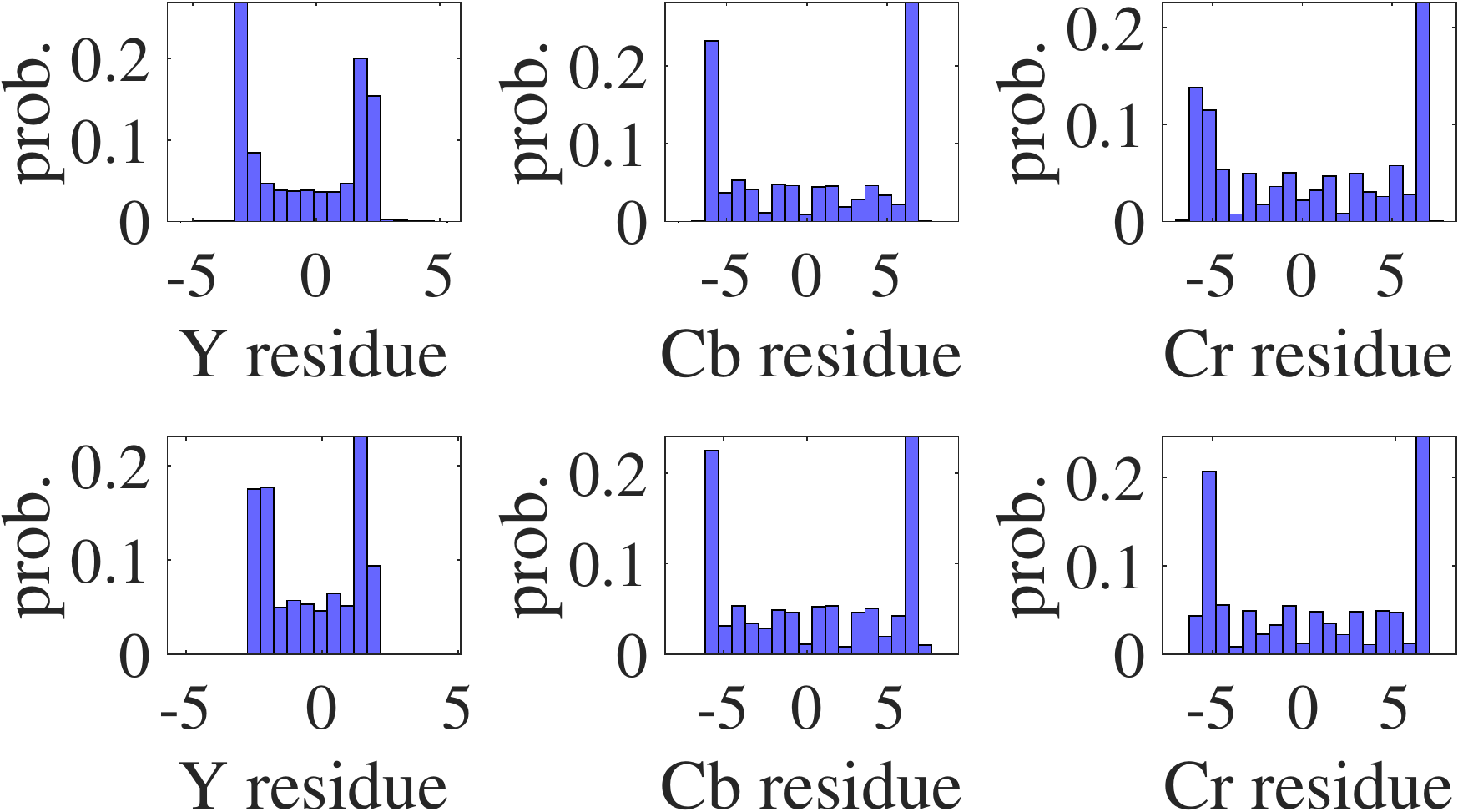} 
\caption{Example perturbation histograms of the proposed method in the $YC_bC_r$ domain on two datasets: $\epsilon^{[Y]}=2.5, \epsilon^{[C_b]}=6, \epsilon^{[C_r]}=6$ on Dataset 1; and $\epsilon^{[Y]}=2, \epsilon^{[C_b]}=6, \epsilon^{[C_r]}=6$ on Dataset 2. The histogram is generated using 5000 adversarial samples of fake face images. }
\label{fig:res_ycc}
\end{figure}

In Fig. \ref{fig:res_ycc}, we show the perturbations generated from the proposed method in the $YC_bC_r$ domain. The $\textrm{1}^{st}$ and $\textrm{2}^{nd}$ rows illustrate the perturbation histograms on Dataset 1 and Dataset 2 with parameters the same as in Table \ref{tab:ASR_fake}. Compared with Fig.\ref{fig:res_fgsm_mim}, perturbations in the $Y$ component approach $\pm 2.5$ on Dataset 1 ($\pm 2$ on Dataset 2). By contrast, perturbations in $C_b$ and $C_r$ components spread away from 0 and concentrate around $\pm 6$. This observation on perturbation residues aligns well with our expectation, which possibly explains the more imperceptible image quality of the proposed attack method. Besides the two datasets as reported, we also experimented on ProGAN \citep{proGAN} and StyleGAN (with image resolution as $512 \times 512$), we have similar conclusion: the proposed method also much improves the visual quality over baseline attacks. Due to page limit, please find more comparison results in the project page.  

\vspace{-5mm} 
\subsection{Comparison on different parameters}
\begin{figure}[h]
\centering
\includegraphics[width=8.8cm, height=4.0cm]{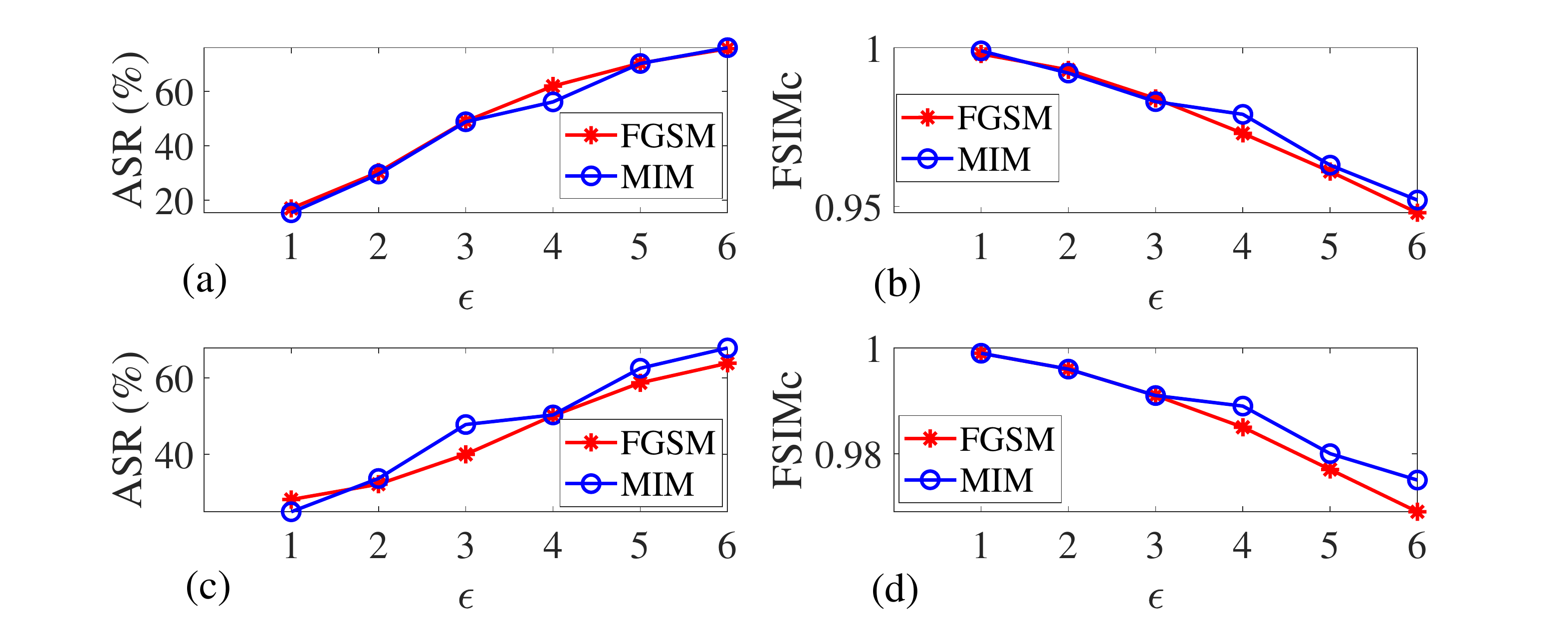} 
\caption{Illustration of the averaged attack success rate and visual quality with different $\epsilon$ values for FGSM and MIM attacks on Dataset 1 ($1^{st}$ row) and Dataset 2 ($2^{nd}$ row). (a) and (c): $ASR^{[p]}$ vs. $\epsilon$; (b) and (d): $\textrm{FSIM}_c$ vs. $\epsilon$.}  
\label{fig:para_sensitivity}
\end{figure}

In Fig.\ref{fig:para_sensitivity}, we show the averaged attack success rates and perceptual quality with different choices of $\epsilon$ for FGSM and MIM attacks on the fake face image subset where the source model is $m_1$. Generally, as the perturbation bound $\epsilon$ increases, $ASR$ increases for both attacks at the cost of visual degradation. To keep comparably high visual quality, e.g., setting $\textrm{FSIM}_c$ as 0.984, the averaged $ASR$ of FGSM and MIM are only about $49.0\%, 46.7\%$. However, the index of the proposed method is $79.2\%$ ($\epsilon^{[c]}=2.5/6/6$) , which is $\textbf{30.2}\%$ and $\textbf{32.5}\%$ higher than FGSM and MIM. Similarly, we can compute the approximate $ASR$ improvement as \textbf{35.1}\% and \textbf{28.5}\% on FGSM and MIM on Dataset 2 when setting $\textrm{FSIM}_c$ as 0.992. This result further convincingly shows the superiority of the proposed method over baseline attacks.

\vspace{-3mm}
\section{Discussion}
\vspace{-2mm}
\label{sec:discussion}
\textbf{Attacking real face images:} Although attacking fake face images poses more threats on forensic models, we also report the attack success rates $ASR^{[n]}$ on the real face image subset (5000 images in total) \cite{styleGAN}. Consistent with the conclusion on fake face images, the proposed method achieves the highest averaged $ASR^{[n]}$ and $\textrm{FSIM}_c$ compared with FGSM and MIM attacks. Interestingly, we observe sharper degradation in attack success rates for each attack method on the real face images subset than those on fake face images. Particularly, the adversarial perturbations may fail completely for the non-deep learning based method \citep{color_disp}. We will further explore this phenomenon in the future work.    

\begin{table}[htb]
\caption{The comparisons of the attack success rate $(\%)$ and visual quality between FGSM, MIM and the proposed method on real face images. The source model is $m_1$. $\epsilon$ is 5.5 for FGSM and 6 for MIM; $\epsilon^{[c]}$ are $4/7/7$ for the proposed method for $Y, C_b, C_r$ channels. The best performances are marked in bold.}
\centering
\begin{adjustbox}{width=0.485\textwidth} 
\begin{tabular}{cccccccccc}
\toprule \noalign{\smallskip}
Attack   & $m_1$ & $m_2$ & $m_3$ & $m_4$ & $m_5$ & $m_6$ & $NDL$ & avg. $ASR^{[n]}$ & $\textrm{FSIM}_c$ \\
\noalign{\smallskip}
\hline
FGSM  & 98.7      & 75.4      & 20.1  & 10.0     & 4.6 & 48.8 & 0    & 41.8  & 0.955     \\ 
\noalign{\smallskip}
\hline
MIM  & 98.7      & 90.8     & 43.5  & \textbf{20.6}     & 7.6  & 81.9  & 0  & 52.2 & 0.955  \\ 
\noalign{\smallskip}
\hline
\textbf{Prop.} & \textbf{98.7} & \textbf{94.4} & \textbf{47.3} & 18.0 & \textbf{11.3} & \textbf{88.9} & 0 & \textbf{53.9} & \textbf{0.965} \\
\noalign{\smallskip}
\bottomrule
\end{tabular}
\end{adjustbox}
\label{tab:ASR_real}
\end{table}

\textbf{Attacks in the $HSV$ domain:} In addition to adversarial attacks in the $YC_bC_r$ domain, we also explored attacks in the $HSV$ domain, since recent study shows relatively large discriminative statistics in the $HSV$ domain for fake face forensics. Our preliminary study shows the inferior performance of $HSV$ than that in the $YC_bC_r$ domain. One possible explanation is the following: There does not exist a clear relationship between $HSV$ channels and the human visual system, thus making it challenging to find adversarial examples both with high attack success rates and imperceptible visual quality.      

\vspace{-3mm}
\section{Conclusion and Future Work}
\label{sec:conclusion}
\vspace{-2mm}
In this work, we studied the imperceptible anti-forensics on GAN-generated fake face imagery detection based on an improved adversarial attack method. For existing attacks, our analysis on perturbation residues shows a significantly reduced perturbation correlation in the $YC_bC_r$ channels when compared with $RGB$ channels, and these perturbations concentrate more on the $Y$ channel than on $C_b$ and $C_r$ channels. Such perturbations can severely degrade the perceptual quality of facial images which have large smooth regions. Thus it makes existing attacks ineffective as a meaningful anti-forensic method. Considering the perception constraint, we propose a novel adversarial attack method that is better suitable for fake face imagery anti-forensics. Specifically, we allocate larger perturbation to $C_b$ and $C_r$ channels which are less sensitive to perception distortion. Simple yet effective, the proposed method achieves both higher adversarial transferability and significant improvement in visual quality when compared with baseline attacks. Moreover, we observe that the proposed method can also fool non-deep learning based forensic detectors with a high attack success rate. This study raises security concerns of existing fake face forensic methods.           

In addition to fake face imagery anti-forensics, we believe all safety-critical forensic models need to be evaluated against such anti-forensics based on adversarial attacks. \textit{More imperceptible} and \textit{transferable}, we hope the proposed anti-forensic algorithm can be a good candidate to evaluate adversarial vulnerability of forensic models. We have released our code to the forensic community for convenient use. In the future, we will further explore the anti-forensic feasibility in related forensic tasks, and develop improved algorithms to counter such anti-forensics.

\vspace{-3mm}
\section*{Acknowledgments}
\vspace{-1mm}
We acknowledge financial support from the Natural Sciences and Engineering Research Council of Canada (NSERC), and Yongwei Wang acknowledges the China Scholarship Council (CSC) for financial support.

\vspace{-3mm}
\bibliographystyle{elsarticle-num}

\bibliography{refs}

\end{document}